\documentclass[conference]{IEEEtran}
\usepackage[english]{babel}

\usepackage{ifpdf}
\usepackage{cite} 
\usepackage{booktabs}
\ifCLASSINFOpdf
\usepackage[pdftex]{graphicx}
\else
\usepackage[dvips]{graphicx}
\fi
\usepackage[cmex10]{amsmath}
\usepackage{algorithmic}
\usepackage{array}
\usepackage{mdwmath}
\usepackage{mdwtab}
\usepackage{url}
\usepackage{pgf}
\usepackage{tikz}
\usepackage{calc}
\usetikzlibrary{matrix, arrows, fit, calc, automata, shapes.gates.logic.US,shapes.gates.logic.IEC,calc, positioning, shapes.geometric, decorations.markings}
\usepackage[latin1]{inputenc}
\usepackage{verbatim}
\usepackage{amsmath,bm,times}
\usepackage{amssymb}
\usepackage{epstopdf}
\usepackage{booktabs}
\usepackage[labelformat=simple]{subcaption}

\DeclareCaptionLabelSeparator{periodspace}{.\quad}
\newcommand{\bftab}{\fontseries{b}\selectfont}

\begin{document}
\title{Block-Matching Optical Flow for Dynamic Vision Sensors: Algorithm and FPGA Implementation}

\author{\IEEEauthorblockN{Min Liu and Tobi Delbruck}
\IEEEauthorblockA{Institute of Neuroinformatics\\
University of Zurich and ETH Zurich,
Zurich, Switzerland\\
Email: minliu@ini.uzh.ch}}
% \and
%\IEEEauthorblockN{Tobi Delbruck}
%\IEEEauthorblockA{Institute of Neuroinformatics\\
%University of Zurich and ETH Zurich\\
%Zurich, Switzerland 8097\\
%Email: tobi@ini.uzh.com}}


%Telephone: (800) 555--1212, Fax: (888) 555--1212}


\maketitle

\begin{abstract}
%\boldmath
Rapid and low power computation of optical flow (OF) is potentially useful in robotics. The dynamic vision sensor (DVS) event camera produces quick and sparse output, and has high dynamic range, but conventional OF algorithms are frame-based and cannot be directly used with event-based cameras. Previous DVS OF methods do not work well with dense textured input and are designed for implementation in logic circuits. This paper proposes a new block-matching based DVS OF
algorithm which is inspired by motion estimation methods used for MPEG video compression. The algorithm was implemented both in software and on FPGA. For each event, it computes the motion direction as one of 9 directions. The speed of the motion is set by the sample interval. Results show that the Average Angular Error can be improved by 30\% compared with previous methods. The OF can be calculated on FPGA with 50\,MHz clock in 0.2\,us per event (11 clock cycles), 20 times faster than a Java software implementation running on a desktop PC. Sample data is shown that the method works on scenes dominated by edges, sparse features, and dense texture. 
\end{abstract}

\IEEEpeerreviewmaketitle

\section{Introduction}
Optical Flow (\textbf{OF}) estimation has always been a core topic in computer vision; it is widely used in segmentation, 3D reconstruction and navigation. It was first studied in the context of neuroscience to understand motion perception in insects and mammals. 
In computer vision, OF describes the motion field induced by camera movement through space. Two well known inexpensive optical flow algorithms are the Lucas-Kanade{\cite{LK OF}} and Horn-Schunck{\cite{HS OF}} methods. 
% Recent years have witnessed active development of OF algorithms based on deep learning technology. Almost all developments are based on camera frames. 
The core of many OF methods is a search over possible flows to select the most likely one at each image or feature location. This search with dense image blocks is expensive and difficult to calculate on an embedded platform in real time.  

The DVS is data-driven rather than regular-sample driven. Regular-sample driven means camera sends the output data at a fixed interval, thus we call it a frame-based camera. However, the DVS output is driven by brightness changes rather than a fixed sample interval. Therefore, new OF methods need to be designed. Benosman et al.~\cite{Benosman Optical Flow} proposed a time-surface method, which combines the 2D events and timestamps into a 3D space and OF is obtained by the local plane fitting. \cite{Benosman 2012} proposed a Lucas-Kanade gradient based method that collects short 2D histograms of events and solves the brightness-constancy constraint on them.  In 2015, Conradt\cite{Conradt embedded OF} proposed a real-time DVS optical flow algorithm implementation on an ARM 7 microcontroller. Barranco~\cite{Barranco 2015} proposed a more expensive phase-based method for high-frequency
texture regions. \cite{Bodo Optical Flow} re-implemented 
 several of these methods in the Java framework jAER~\cite{jAER website} and compared them with the earliest jAER method based on time-of-flight of oriented edges. Its conclusion was that 
all methods offered comparable accuracy for sharp and sparse edges, but all fail on textured or low spatial frequency inputs, because the underlying assumptions (e.g. smooth gradients or isolated edges) are violated. This paper also introduced the use of an integrated 
camera inertial measurement unit~(\textbf{IMU}) to obtain 
ground truth global optical flow from camera rotation and published a benchmark dataset from a 240x180 pixel DVS camera, which we use here. Most of the existing work is based on PC software algorithms\cite{Benosman Optical Flow} \cite{Benosman 2012}~\cite{Barranco 2015}~\cite{Bodo Optical Flow}. Though \cite{Conradt embedded OF} 
is based on an embedded system and can work in real-time, it was only characterized for camera rotations, not camera translation through space, and its use of direct time of flight of events makes it unlikely to work well with dense textured scenes and to suffer from aperture problems for edges. 

In video technology, OF is called motion estimation (\textbf{ME}) and is widely used in exploiting the temporal redundancy of video sequences for video compression standards, such as MPEG-4 and H.263{\cite{Shahrukh Agha ME}}. The pipeline for ME includes block matching. Block matching means that rectangular blocks of pixels are matched between frames to find the best match. Block matching is computationally expensive. That is why it is now widely implemented in dedicated logic circuits. 
In order to address this problem, an example of logic ME implementation based on block matching is presented in Shahrukh{\cite{Shahrukh Agha ME}}. Our paper proposes an event-based block matching algorithm to calculate OF on FPGA. 

The paper is organized as follows: Section II introduces our system architecture and algorithm, section III shows experimental results, and Section IV concludes the paper.

\section{Proposed Method}
The output of DVS is a stream of brightness change events. Each event has a microsecond timestamp, a pixel address, and a binary polarity describing the sign of the brightness change. Each event signifies a change in brightness of about 15\% since the last event from the pixel. In this work, events are accumulated into time slice frames as binary images ignoring the polarity, since our aim is for minimum logic and memory size. Here we will refer to these bitmap frames as slices. 

A block is a square centered around the incoming event's location. Matching is based on a distance metric. In our work, we implemented Hamming Distance (\textbf{HD}) as the distance metric. HD is the count of the number of differing bits. For bitmaps, HD is exactly the same as the better-known Sum-of-Absolute-Differences (SAD).

The software implementation is open source. It is called PatchMatchFlow\cite{PatchMatchFlow Source Code} in jAER.

\subsection{System Evaluation Architecture}
The hardware evaluation system is divided into two parts, one for data sequencing and monitoring and the other for the algorithm implementation. For the first part, we use a monitor-sequencer board \cite{Sequence board 2007} designed by the Univ. of Seville. The sequencer converts the event-based benchmark dataset\cite{Bodo Optical Flow} into real-time hardware events sent to the OF FPGA. During OF calculation, the monitor collects the OF events and sends them over USB to jAER for rendering and analysis. In this way we can compare software and hardware processing of the OF algorithm.  In this work, we only used prerecorded data to allow systematic comparison between software and hardware implementations.

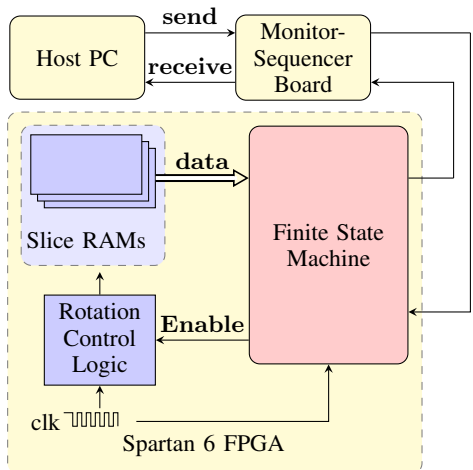
\begin{figure}[!t]
    \centering
    \begin{comment}
        Title: The Whole System Architecture
        You can put the comments about the figure here.
    \end{comment}

    \newcommand{\mx}[1]{\mathbf{\bm{#1}}} % Matrix command
    \newcommand{\vc}[1]{\mathbf{\bm{#1}}} % Vector command
\pagestyle{empty}

    % We need layers to draw the block diagram
    \pgfdeclarelayer{background}
    \pgfdeclarelayer{foreground}
    \pgfsetlayers{background,main,foreground}

    % Define a few styles and constants
    \tikzstyle{sensor}=[draw, fill=blue!20, text width=4em, 
    text centered, minimum height=2.5em]
    \tikzstyle{ann} = [above, text width=5em]
    \tikzstyle{naveqs} = [sensor, text width=6em, fill=red!20, 
    minimum height=10em, rounded corners]
    \tikzstyle{general} = [sensor, text width=5em, fill=yellow!20, 
    minimum height=3.5em, rounded corners]
    \tikzstyle{vecArrow} = [thick, decoration={markings,mark=at position 1 with {\arrow[semithick]{open triangle 60}}},
        double distance=1.4pt, shorten >= 5.5pt,
        preaction = {decorate},
        postaction = {draw,line width=1.4pt, white,shorten >= 4.5pt}]
    \tikzstyle{innerWhite} = [semithick, white,line width=1.4pt, shorten >= 4.5pt]

    \def\blockdist{3.3}
    \def\edgedist{2.5}
    \begin{tikzpicture}[scale = 0.6]
    	\small
        \node (naveq) [naveqs] {Finite State Machine};
        % Note the use of \path instead of \node at ... below. 
        \path (naveq.140)+(-\blockdist,0) node (sliceRam1) [sensor] {};
        \path (naveq.140)+(-\blockdist-0.15,0.15) node (sliceRam2) [sensor] {};
        \path (naveq.140)+(-\blockdist-0.3,0.3) node (sliceRam3) [sensor] {};
        %\path (naveq.-130)+(-\blockdist,0) node (rotation) [sensor] {Rotation Control Logic};
        \node at ($(naveq.-130)+(-\blockdist,0)$) (rotation) [sensor] {Rotation Control Logic};

        \path (sliceRam1.north) + (-0.5, 2) node (host) [general] {Host PC};
        \path (host.east) + (3.5, 0) node (seq) [general] {Monitor-Sequencer Board};

        \path [draw, -stealth] (host.20) -- node [above] {$\vc{send}$} (seq.160);
        \path [draw, -stealth] (seq.200) -- node [above] {$\vc{receive}$} (host.-20);

        \path [draw] (seq.20) -- ([xshift = 2.2cm]seq.20);
        \path [draw, -stealth] ([xshift = 2.2cm]seq.20) |- node [above] {} (naveq.-40);

        \path [draw] (naveq.40) -- ([xshift = 1cm]naveq.40);
        \path [draw, -stealth] ([xshift = 1cm]naveq.40) |- node [above] {} (seq.-20);

        % Unfortunately we cant use the convenient \path (fromnode) -- (tonode) 
        % syntax here. This is because TikZ draws the path from the node centers
        % and clip the path at the node boundaries. We want horizontal lines, but
        % the sensor and naveq blocks aren't aligned horizontally. Instead we use
        % the line intersection syntax |- to calculate the correct coordinate
        \path [draw, vecArrow] (sliceRam1) -- node [above] {$\vc{data}$} 
        (naveq.west |- sliceRam1) ;
        % We could simply have written (sliceRam) .. (naveq.140). However, it's
        % best to avoid hard coding coordinates
        \path [draw, stealth-] (rotation) -- node [above] {$\vc{Enable}$} 
        (naveq.west |- rotation);
        \node (RAMs) [below of=sliceRam3] {Slice RAMs};
        \path (naveq.south west)+(-1, -1.8) node (INS) {Spartan 6 FPGA};

        % Now it's time to draw the colored IMU and INS rectangles.
        % To draw them behind the blocks we use pgf layers. This way we  
        % can use the above block coordinates to place the backgrounds   
        \begin{pgfonlayer}{background}
            % Compute a few helper coordinates
            \path (sliceRam3.west |- naveq.north)+(-0.5,0.3) node (a) {};
            \path (INS.south -| naveq.east)+(+0.3,-0.2) node (b) {};
            \path[fill=yellow!20,rounded corners, draw=black!50, dashed]
            (a) rectangle (b);

            \path (sliceRam3.north west)+(-0.2,0.2) node (a) {};
            \path (RAMs.south -| sliceRam1.east)+(+0.2,-0.2) node (b) {};
            \path [fill=blue!10,rounded corners, draw=black!50, dashed]
            (a) rectangle (b) node (ramRect) {};

            \draw [-stealth] (rotation.north) -- ([yshift=0.5cm]rotation.north);   
            %% This one is for generating the random period square waves. 
%            \draw (0,0) --(1,0)|- ++(rnd,1)
%            \foreach \x in {1,...,10}{-|++(rnd,-1) -| ++(rnd,1)}
%            -| ++(rnd,-1);

            \tikzset{clock/.style={ append after command={%
                        \pgfextra
                        \draw ($(rotation.south)+(-0.8,-0.6)$) --++(0.1,0)
                        \foreach \x in {1,...,5}{-|++(0.1,-0.3) -| ++(0.1,0.3)}
                        -| ++(0.1,-0.3);
            \endpgfextra}}}

            \node [clock] at ($(rotation.south)+(-1.2,-0.8)$) (clk) {clk}; 
            \draw[-stealth] ([xshift=1.5cm, yshift=0.2]clk.east) -| (naveq.south);
            \draw[-stealth] ([yshift=-0.5cm]rotation.south) -- (rotation.south);

        \end{pgfonlayer}
    \end{tikzpicture}
    \caption{System Architecture}
    \label{fig1}
\end{figure}

\begin{figure}[!t]
    \centering
    \begin{tikzpicture}[->,>=stealth',shorten >=1pt,auto,node distance=3cm, 
        semithick]
        \tikzstyle{every state}=[fill=blue, draw=none, text=white, align=center, minimum size = 2cm, text width = 1.5cm, scale = 0.55];  % Set the nodes' style
        \tikzstyle{every edge}=[draw, thick];     % Set the edges' style

        \node[initial,state] (A)                    {IDLE};
        \node[state] (B) [right of = A]                    {Read};
        \node[state]         (C) [right of=B] {Data Check};
        \node[state]         (D) [right of=C] {Extract Events};
        \node[state]         (E) [below of=D] {Read Blocks};
        \node[state]         (F) [below of=E]       {SAD /HD};
        \node[state]         (G) [left of=F]       {Get Minimum};
        \node[state]         (H) [left of=G]       {Send data};
        \node[state]         (I) [below of=A]       {Timeout Check};
        \node[state]         (J) [below of=I]       {RAM Rotation};

        \path (A) edge[bend left]              node[above] {req = 0} (B)
        (A) edge [loop above] node {req = 1} (A)
        (B) edge [bend left]          node {} (C)
        (C) edge [bend left] node {No} (A)
        edge     [bend left]  node {Yes} (D)
        (D) edge     [bend left]  node {} (E)
        (E) edge     [bend left]  node {} (F)
        (F) edge     [bend left]  node {} (G)
        (G) edge     [bend left]  node {} (H)
        (H) edge     [loop below]  node {ack = 1} (H)
        edge     [above right]  node {ack = 0} (I)
        (I) edge     [bend right]  node {No} (J)
        edge     [bend left]  node {Yes} (A);
    \end{tikzpicture}
    \caption{Finite state machine}
    \label{fig2:FSM}
\end{figure}
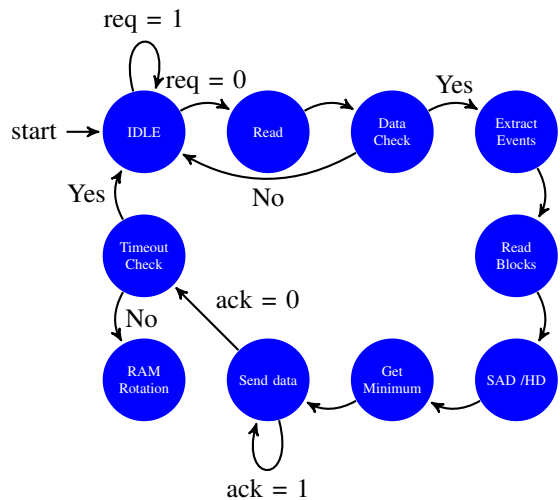

The OF architecture (Fig \ref{fig1}) contains three main modules: the finite state machine (\textbf{FSM}), random access memory block memory (\textbf{RAMs}) and rotation control logic. The architecture of the FSM is shown in Fig \ref{fig2:FSM}. The FSM consists of three parts: data receiving module, OF calculation module, and data sending module. The data sending and data receiving module communicate with the monitor-sequencer.   The OF
 module is described in the section \ref{Optical Flow algorithms}.

Three 240x180-pixel DVS event bitmap slices are stored in RAM. These slices are like binary image frames from conventional cameras but in the case of DVS we can arbitrarily select the slice interval. One is the current collecting slice starting at time \textit{t} and the 
other two are the past two slices starting 
at times \textit{t-d} and \textit{t-2d}. 
\textit{d} is the slice duration. %in this paper we used d=100\,ms to match the slow motion in the dataset.
At intervals of \textit{d}, the rotation control logic rotates the three slices. 
The \textit{t} slice accumulates new data. 
It starts out empty 
and gradually accumulates events, so it cannot be used for matching to past slices.
The two past slices are used for OF, but the OF computation is done at the location of each event stored into the \textit{t} slice, and thus is driven by these events. 
Slices are stored in  
block RAM on the FPGA. 
The total size of the RAM is 240x180x3, matching the DVS pixel array size. 
It is generated by the IP Core of Xilinx. 

\subsection{Optical Flow algorithm} \label{Optical Flow algorithms}

When an event arrives, a single reference block from slice \textit{t-d} and 9 blocks from slice \textit{t-2d} are sent to the HD module to calculate the distances. In the current implementation, the block contains 9x9 pixels. For the \textit{t-d} slice, we use only one center block as the reference.  The algorithm finds the most similar block on the \textit{t-2d} slice. According to the brightness-constancy assumption of OF, we should see a similar block in the \textit{t-2d} slice for the block that best matches the actual OF. We search over the 8 blocks centered on the 8 neighbors of the
current event address and one block centered on the reference and choose the one with minimum distance. % (Zero motion is not checked in the current implementation.)

% \subsubsection{Sum of Absolute Distance}
% SAD is widely used in video compression motion estimation because it is computationally cheap. An SAD implementation on FPGA was introduced in \cite{Wong sad on FPGA}.  In our implementation we use 9x9=81 absolute difference circuits (Fig~\ref{fig:sad}) to compute SAD in 1 clock cycle for each block comparison. 

\subsubsection{Hamming Distance}
The implementation of one HD block is shown in
Fig~\ref{fig:hmd}.  A total of 81 XOR logic gates receive input from corresponding pixels on the slices. The XOR outputs are summed to compute the HD. 

\subsubsection{Minimum Distance Computation}
The last step of the algorithm is to find the minimum distance candidate. Part of the novel minimum circuit is shown in Fig \ref{fig:sort}. It is a parallel implementation that outputs the index of the minimum distance direction. For instance, if we need to find the minimum among 5 data: HD0-4 (output from Fig \ref{fig:hmd}), the circuit can be divided into 5 parts. The first part in Fig \ref{fig:sort} compares HD0 with all the other data and outputs a count of how many times data0 is larger than HD1-4.  The other 4 parts are implemented in the same way and all those parts are computed concurrently. At the end, the part whose sum is zero is the minimum candidate. Thus the minimum distance candidate is determined in one clock cycle. 

\begin{figure}[!t]
    \centering
    \tikzstyle{branch}=[fill,shape=circle,minimum size=3pt,inner sep=0pt]
    \begin{tikzpicture}[label distance=2mm, scale = 0.7]

        \node (x5) at (0,8) {$\rm block_{t-d}[0]$};
        \node (x6) at (0,7.4) {$\rm block_{t-2d}[0]$};
        \node (x7) at (0,6.4) {$\rm block_{t-d}[1]$};
        \node (x8) at (0,5.8) {$\rm block_{t-2d}[1]$};
        \node (x9) at (0,4.8) {$\rm block_{t-d}[79]$};
        \node (x10) at (0,4.2) {$\rm block_{t-2d}[79]$};
        \node (x11) at (0,3.2) {$\rm block_{t-d}[80]$};
        \node (x12) at (0,2.6) {$\rm block_{t-2d}[80]$};

        \node[xor gate US, draw, logic gate inputs=nn] at ($(x5) + (4, -0.5)$) (Xor1) {};
        \node[xor gate US, draw, logic gate inputs=nn] at ($(x7) + (4, -0.5)$) (Xor2) {};
        \node[xor gate US, draw, logic gate inputs=nn] at ($(x9) + (4, -0.5)$) (Xorn-1) {};
        \node[xor gate US, draw, logic gate inputs=nn] at ($(x11) + (4, -0.5)$) (Xorn) {};

        \draw (x5) -- ($(x5)+(2,0)$) |-  (Xor1.input 1);
        \draw (x6) -- ($(x6)+(2,0)$) |-  (Xor1.input 2);

        \draw (x7) -- ($(x7)+(2,0)$) |-  (Xor2.input 1);
        \draw (x8) -- ($(x8)+(2,0)$) |-  (Xor2.input 2);

        \draw (x9) -- ($(x9)+(2,0)$) |-  (Xorn-1.input 1);
        \draw (x10) -- ($(x10)+(2,0)$) |-  (Xorn-1.input 2);

        \draw (x11) -- ($(x11)+(2,0)$) |-  (Xorn.input 1);
        \draw (x12) -- ($(x12)+(2,0)$) |-  (Xorn.input 2);

        \path (x8) -- node[auto=false]{\ldots} (x9);

        \draw (8,2.5)coordinate (O)--++(30:1)coordinate (A)--++(90:4)coordinate (B)--++(150:1)coordinate (C)--cycle;
        \draw ($(A)!0.5!(B)$)--++(0:1)node[right]{$\rm HD$};
        %\draw ($(O)!0.5!(A)$)--++(-90:1)--++(180:2)node[left]{$b$};
        %\draw ($(O)!0.25!(A)$)--++(-90:0.5)--++(180:1.75)node[left]{$a$};
        %\draw ($(O)!0.75!(A)$)--++(-90:1.5)--++(180:2.25)node[left]{$c$};
        \foreach \y/\t in {0.1/1,0.2/2,0.7/n-1,0.8/n} {
            \draw ($(C)! \y*1.1 !(O)$)--++(180:1) node[above] {$x_{\t}$};
            \draw (Xor\t.output) -- ([xshift=0.5cm]Xor\t.output) |- ($(C)! \y*1.1 !(O)$);
        }
        \node[] at (8.5,5) {+};           
    \end{tikzpicture}
    \caption{Hamming Distance implementation for one 9x9 block match. There are 9 of these circuits for the 9 flow directions.}
    \label{fig:hmd}
\end{figure}
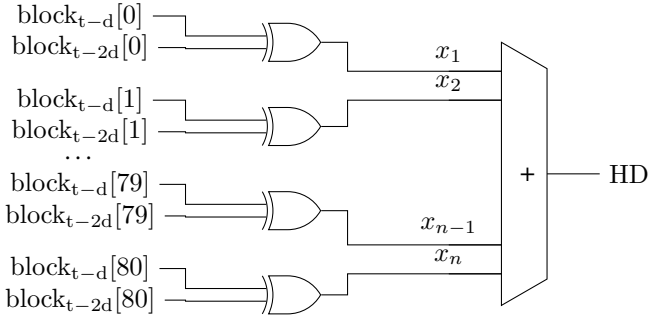

\begin{figure}[!t]
    \centering
    \tikzstyle{branch}=[fill,shape=circle,minimum size=3pt,inner sep=0pt]
\begin{tikzpicture}[label distance=2mm, scale = 0.6]    

	\node (squa1) at (4,2) [draw,thick,minimum width=1cm,minimum height=1cm] {$\geq$};
    \node (squa2) [below = 0.2cm of squa1] [draw,thick,minimum width=1cm,minimum height=1cm] {$\geq$};
    \node (squa3) [below = 0.2cm of squa2] [draw,thick,minimum width=1cm,minimum height=1cm] {$\geq$};    
    \node (squa4) [below = 0.2cm of squa3] [draw,thick,minimum width=1cm,minimum height=1cm] {$\geq$};
    
    \node (data1) at ($(squa1.210) + (-4, 0)$) {HD0};
    \node (data2) at ($(squa1.150) + (-4, 0)$) {HD1};
    \node (data3) at ($(squa2.150) + (-4, 0)$) {HD2};
    \node (data4) at ($(squa3.150) + (-4, 0)$) {HD3};
    \node (data5) at ($(squa4.150) + (-4, 0)$) {HD4};    
    
    \draw[-stealth] (data1) -- (squa1.210);
    \draw[-stealth] (data2) -- (squa1.150);
    \draw[-stealth] (data3) -- (squa2.150);
    \draw[-stealth] (data4) -- (squa3.150);
    \draw[-stealth] (data5) -- (squa4.150);

	\draw[-stealth] ($(squa1.210) + (-2, 0)$) node[branch]{} |- (squa2.210);
	\draw[-stealth] ($(squa2.210) + (-2, 0)$) node[branch]{} |- (squa3.210);
	\draw[-stealth] ($(squa3.210) + (-2, 0)$) node[branch]{} |- (squa4.210);
	\draw[-stealth] ($(squa4.210) + (-2, 0)$) node[branch]{} |- (squa4.210);
       
    % This is the adder shape.
    \draw (7,-4)coordinate (O)--++(30:1)coordinate (A)--++(90:5)coordinate (B)--++(150:1)coordinate (C)--cycle;
    \draw ($(A)!0.5!(B)$)--++(0:1)node[right, text width=2cm]{\# of inputs smaller than HD0};

    \foreach \y/\t in {0.1/1,0.2/2,0.7/3,0.8/4} {
    \draw ($(C)! \y*1.1 !(O)$)--++(180:1) node[above] {};
    \draw (squa\t.east) -- ([xshift=0.5cm]squa\t.east) |- ($(C)! \y*1.1 !(O)$);
    }
    \node[] at (7.5,-1) {+};           

    \end{tikzpicture}
    \caption{Sort algorithm implementation block for HD0, simplified for 5 inputs rather than 9. There are 9 of these blocks.}
    \label{fig:sort}
\end{figure}
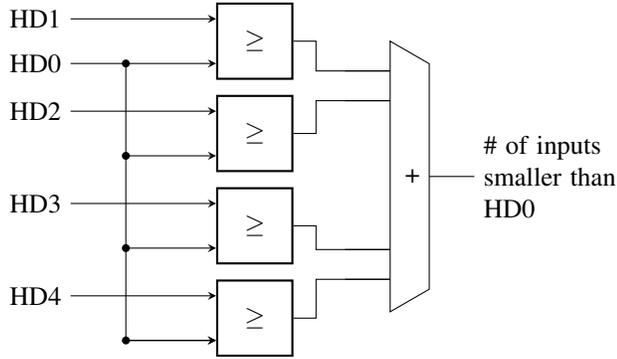

\section{Experimental Results}
We used the Xilinx Spartan 6 family chip xc6slx150t to implement our algorithm. It has 184304 Flip-Flops and 92152 LUTs and 4MB block memory. The implemented OF design occupies 0.9\% of the Flip-Flops, 5\% of the LUTs and 5\% of the block RAM. For the test dataset, we use the event-based optical flow benchmark dataset in \cite{Bodo Optical Flow} which also provides the evaluation method and the ground truth.  

\begin{figure}
    \centering
%     \begin{subfigure}[b]{0.2\textwidth}
%       	\centering
%         \includegraphics[width=1.3in]{pics/barTrans_HDist_small.JPG}
%         \caption{Bar translation}
%         \label{fig:bar}
%     \end{subfigure}
%     ~ %add desired spacing between images, e. g. ~, \quad, \qquad, \hfill etc. 
%       %(or a blank line to force the subfigure onto a new line)
%     \begin{subfigure}[b]{0.2\textwidth}
%       	\centering
%         \includegraphics[width=1.25in]{pics/transSquare_PC_small.JPG}
%         \caption{Square translation}
%         \label{fig:square}
%     \end{subfigure}
    ~ %add desired spacing between images, e. g. ~, \quad, \qquad, \hfill etc. 
    \begin{subfigure}[b]{0.3\textwidth}
        \begin{center}
            \includegraphics[width=2in]{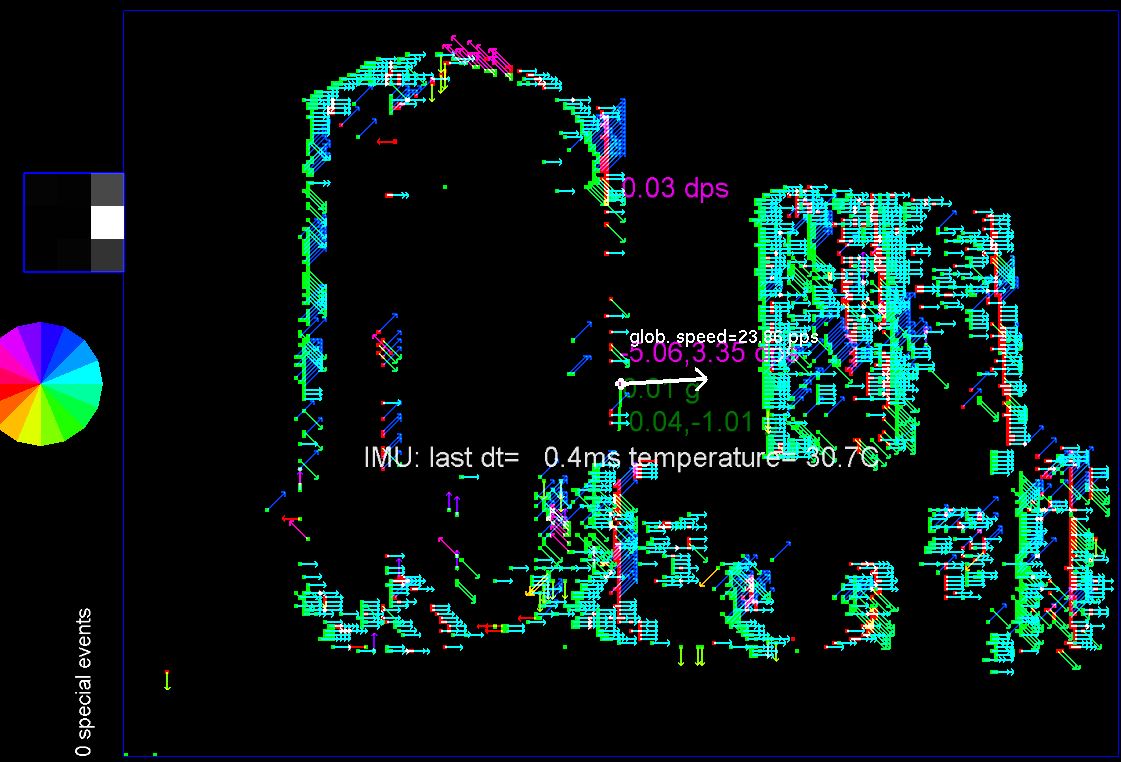}
        \end{center}
        \caption{Boxes translation}
        \label{fig:boxes}
    \end{subfigure}
    \begin{subfigure}[b]{0.3\textwidth}
        \begin{center}
            \includegraphics[width=2in]{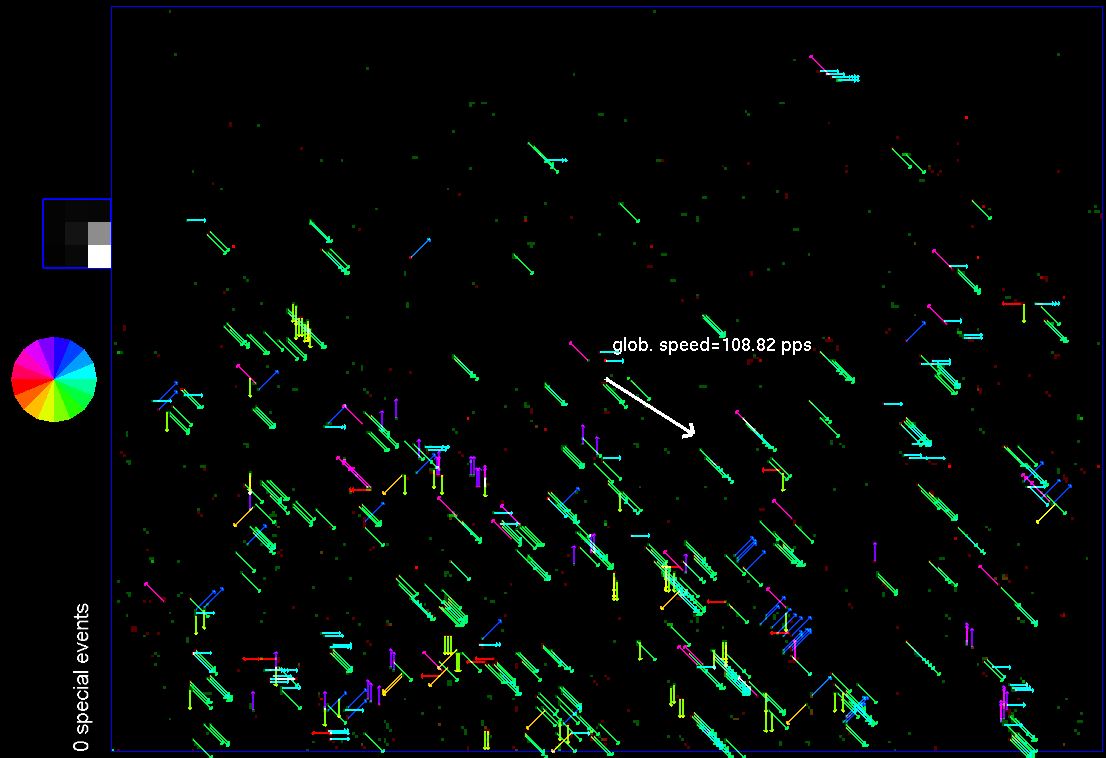}
        \end{center}
        \caption{Pavement on grass}
        \label{fig:grassPavement}
   \end{subfigure}
   \begin{subfigure}[b]{0.3\textwidth}
        \begin{center}
            \includegraphics[width=2in]{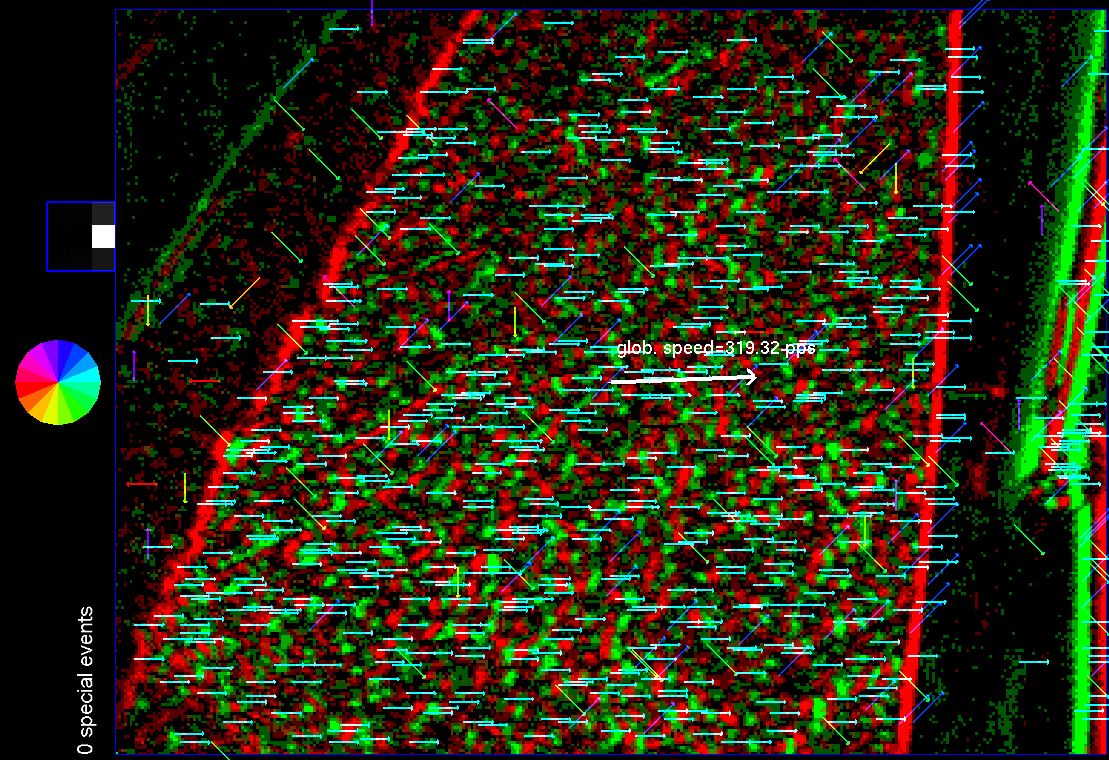}
        \end{center}
        \caption{Gravel}
        \label{fig:grassPavement}
   \end{subfigure}
    \caption{OF Results. The arrows are the flow vectors and their length represents the speed (determined by the slice duration \textit{d}). DVS On events are green and Off events are red.  The color wheel indicates the flow vector direction color. The 2D gray scale histogram above each color wheel shows the distribution of flow event directions (here we use 9 direction bins) in the time slice. The brightest bin votes the highly possible direction of the global motion. % Global speed estimation is also displayed. 
% (a) and (b) are from the synthetic dataset; a is the bar translation, b is the square translation. 
(a) is the boxes scene from~\cite{Bodo Optical Flow} with $d=40{\rm ms}$. (b) is pavement recorded by a down-looking DVS; $d=10{\rm ms}$. (c) is
   a gravel area with $d=3{\rm ms}$. For clarity, downsampling was used to compute 1 flow event every 100 DVS events.}
   \label{fig:result}
\end{figure}
We tested three sample datasets. % and two of them are synthesized data which are just used for algorithm correction checking are not shown here. %The other two 
All of them are real DVS data: the box translation, pavement, and gravel corresponding to edge, sparse points, and dense texture respectively.  
% The bar translation and the square translation are both synthesized data, but the box translation and pavement on grass field are the real data from the camera. The bar is moving from left to right and the square is translating from left-bottom to right-top.
The boxes scene has a box in the foreground and clutter in the background and the camera pans to the left, producing rightwards global translation mostly of extended edges. In the pavement dataset, the camera was down-looking and carried by hand; the flow points downwards and to the right. Imperfections in the pavement cause sparse features. The gravel dataset is recorded outside and has dense texture; movement is eastward.

The block-matching OF results are shown in Fig \ref{fig:result}. %For the bar translation, the OF vectors at the top of the bar are correct (they point east) because there is no motion ambiguity. But at the bottom of the bar, it is impossible to determine the motion non-ambiguously, since multiple directions result in the same distance. For the square box example, most of the vectors point in the correct direction (northeast). 
It can be seen that in each scene, most vectors point correctly east for box translation, southeast for the pavement scene, and east for the gravel scene. Errors are mostly caused by DVS noise or aperture ambiguity for the extended edges.

\subsection{Accuracy analysis}
\cite{Bodo Optical Flow} proposed two ways to calculate event-based OF accuracy, based on similar metrics used for conventional OF. One is called Average Endpoint Error (AEE) and the other is Average Angular Error (AAE). AAE measures error in the direction of estimated flow and AEE includes speed error. These two methods are already implemented in jAER{\cite{jAER website}. They use IMU data from a pure camera rotation along  with lens focal length as the ground truth. Since the output data of the sequencer lacks IMU data, we measured the OF accuracy using the PC implementation. The algorithm pipeline between FPGA and PC is identical, so it will not
influence the accuracy. The result is also compared with \cite{Bodo Optical Flow}. We chose two variants of the event-based Lucas-Kanade and Local Plane algorithms.
The errors from all the algorithms are shown in Table \ref{tab:error}. $PM_{hd}$ represents the block matching algorithm with HD metric. % $LK_{sg}$ and $LK_{bd}$ are two variants of Lucas-Kanade. $LP_{orig}$ and $LP_{sg}$ represents two variants of the Local Planes algorithm. 

% This is for the table aligning in two rows.
% \begin{table}[htbp]
%  \centering
%  \caption{AAE comparison}\label{tab:aae}
%  \begin{tabular}{lcl}
%   \toprule
%   \bftab AAE & \bftab transBoxes & \bftab grassPavement \\
%   \midrule
%       % $PM_{sad}$ & 44.56$\pm$37.15 & 0$\pm$0 \\
%       $PM_{hd}$ & \bftab 42.68$\pm$33.82 & \bftab 56.19$\pm$39.99 \\
%       %$LK_{sg}$ & \bftab 30.30$\pm$44.35 & \bftab 50.11$\pm$29.59 \\
%       $LK$ &  98.92$\pm$42.24 & 94.93$\pm$39.42 \\
%       $LP_{orig}$ & 77.18$\pm$33.73 & 85.23$\pm$38.97 \\
%       $LP_{sg}$ & 47.52$\pm$54.44 & 80.95$\pm$40.96 \\
%   \bottomrule
%  \end{tabular}
% \end{table}

% \begin{table}[htbp]
% \centering
% \caption{AEE comparison}\label{tab:aee}
%  \begin{tabular}{lcl}
%   \toprule
%   \bftab AEE & \bftab transBoxes & \bftab grassPavement \\
%   \midrule
%       % $PM_{sad}$ & \bftab 18.30$\pm$7.02 & 0$\pm$0 \\
%       $PM_{hd}$ & \bftab 17.86$\pm$6.31 & \bftab 128.67$\pm$56.59 \\
%       %$LK_{sg}$  & 24.72$\pm$26.11 & 1218.69$\pm$6736.27 \\
%       $LK$  & 37.00$\pm$15.18 & 134.69$\pm$64.76 \\
%       $LP_{orig}$ & 93.02$\pm$107.02 & 302.89$\pm$294.82 \\
%       $LP_{sg}$  & 98.32$\pm$82.5 & 287.59$\pm$192.90 \\
%   \bottomrule
%  \end{tabular}
% \end{table}

% This is for tables aligning in the same row.
\begin{table}
\caption{OF algorithm's accuracy}\label{tab:error}
\begin{subtable}{.5\linewidth}\centering
{\begin{tabular}{ccc}
      \toprule
      \bftab AAE & \bftab transBoxes \\ % \bftab grassPavement \\ %& transSquare \\
      \midrule
      %$PM_{sad}$ & 44.56$\pm$37.15 \\ %& 0$\pm$0 \\
      $PM_{hd}$ & 42.68$\pm$33.82 \\ %& 0$\pm$0 \\
      $LK_{sg}$ & \bftab 30.30$\pm$44.35 \\ %& 0$\pm$8 \\
      $LK_{bd}$ &  98.92$\pm$42.24 \\ %& 0$\pm$0 \\
      $LP_{orig}$ & 77.18$\pm$33.73 \\%& 0$\pm$0 \\
      $LP_{sg}$ & 47.52$\pm$54.44 \\ %& 0$\pm$0 \\
      \bottomrule  
\end{tabular}}
\caption{AAE comparison}\label{tab:aae}
\end{subtable}%
\begin{subtable}{.5\linewidth}\centering
{\begin{tabular}{c c c}
      \toprule
      \bftab AEE & \bftab transBoxes \\ %& transSquare \\
      \midrule
      %$PM_{sad}$ & \bftab 18.30$\pm$7.02 \\ %& 0$\pm$0 \\
      $PM_{hd}$ & \bftab 17.86$\pm$6.31 \\ %& 0$\pm$0 \\
      $LK_{sg}$  & 24.72$\pm$26.11 \\ %& 0$\pm$0 \\
      $LK_{bd}$  & 37.00$\pm$15.18 \\ %& 0$\pm$0 \\
      $LP_{orig}$ & 93.02$\pm$107.02 \\ %& 0$\pm$0 \\
      $LP_{sg}$  & 98.32$\pm$82.5 \\ %& 0$\pm$0 \\
      \bottomrule
\end{tabular}}
\caption{AEE comparison}\label{tab:aee}
\end{subtable}
\end{table}

As shown in Table \ref{tab:error}, the block matching algorithm has the best accuracy for AEE and second-best for AAE, partly from an appropriate choice of the sample rate that matches the dataset motion. 

Fig \ref{fig:patchAAE} shows the relationship between the block radius and AAE. It indicates that bigger block dimension leads to better accuracy. However, larger blocks consume more logic and reduce spatial resolution of the flow. The comparison between PC and FPGA implementation complexity is discussed next, in \ref{Time Analysis}.

\begin{figure}[!t]
\centering
\includegraphics[width=2.5in, scale = 2]{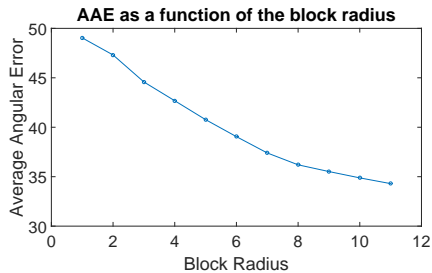}
\caption{The relationship between the block radius and AAE}
\label{fig:patchAAE}
\end{figure}

\subsection{Time complexity analysis} \label{Time Analysis}
The time complexity of the software grows quadratically with the block size while only linearly in FPGA. The processing time of the algorithm contains three parts: reading data from three slices, HD calculation and looking for the minimum. Both FPGA implementation and software implementation on PC consume linear time to read data from RAM since multiple data cannot be read from one RAM simultaneously. However, the latter two parts take constant time (2 clock cycles) on FPGA while quadratic time on PC. In summary, processing time on FPGA is (block dimension + 2) cycles. In this paper, FPGA runs at 50MHz frequency and the block dimension is 9. Thus the whole algorithm will take only 220ns per event, i.e. 0.22us. On PC, it takes 4.5us per event for (admittedly non-optimized) jAER to run the algorithm. The implementation on FPGA is 20 times faster than that on the PC. The current implementation uses single port RAM and could be further sped up by using multiple banks.

\section{Conclusion}
In this paper, we proposed a new method to estimate the event-based optical flow on FPGA in real time. The software computational cost of Hamming Distance increases quadratically as the block size increases, however, in FPGA, all the bits in the block can be calculated at the same time which leads  to a constant time for all block sizes. This greatly reduces the overall computation time for the FPGA implementation, which is 20 times faster than the software implementation. In the current implementation, every single incoming event is processed (allowing an input event rate of up to 5\,Meps to be handled using a modest FPGA clock of only 50\,MHz). However, processing every event is not required, as illustrated in Fig.~\ref{fig:grassPavement}, where OF computation is downsampled, but the DVS events still indicate locations to estimate the flow.

There are three possible improvements. The current implementation estimates only direction of flow and not speed. Measuring speed requires additional search distances and there are well-known algorithms for efficient search~\cite{BMA}.  Secondly, other distance metrics should be explored because event sequences collected onto the slices usually have different length due to noise and HD is somewhat ambiguous~\cite{Weighted HD}. Finally, we will implement feedback control on the slice duration to better exploit the unique feature of DVS event output that it can be processed at any desired sample rate. %This capability is a key distinguishing characteristic from frame-based vision, where sample rate and processing rate are inextricably coupled. It could allow a block-matching approach for DVS that achieves high OF accuracy even with only small search distances and modest hardware resources.


\section*{Acknowledgment}

Funded by Swiss National Center of Competence in Research Robotics (NCCR Robotics). We thank H. Liu and the Architecture and Computer Technology group, Univ. of Seville, Spain for support with testing.



%\bibliographystyle{IEEEtran}

\end{document}